\RequirePackage[svgnames]{xcolor}

\documentclass[11pt,letterpaper]{mystyle}
\usepackage{lmodern}
\usepackage{tikz}
\usetikzlibrary{
  shapes,               % 矩形、圆形等节点形状
  arrows.meta,          % 各种箭头样式
  positioning,          % 方便 relative 定位
  decorations.pathreplacing, % 括号、波浪线等装饰
  fit                   % 用于 \node[fit=…] 包装一组节点
}

\usepackage[all]{hypcap}
\usepackage[svgnames]{xcolor}
\usepackage[comma,authoryear,compress]{natbib}
\usepackage{hyperref}[citecolor=lightblue]

\hypersetup{
    colorlinks = true,
    citecolor = {YaleBlue},
}

\usepackage{algorithm}
\usepackage{algorithmicx}
\usepackage{algpseudocode}
\usepackage{caption} 
\usepackage{microtype}
\usepackage{graphicx}
\expandafter\def\csname ver@subfig.sty\endcsname{}
\usepackage{booktabs} %
\usepackage{float}
\usepackage{bigstrut}
\usepackage{booktabs}
\usepackage{amsmath}
\usepackage{amssymb}
\usepackage{mathtools}
\usepackage{amsthm}
\usepackage{mathrsfs}
\usepackage{nicefrac}
\usepackage{dsfont}
\usepackage{enumitem}
\usepackage{subcaption}
\usepackage{graphicx,subfig}
\usepackage{cleveref}
\usepackage{bxcoloremoji}
\usepackage{epigraph}
\usepackage{float}

\usepackage{arydshln}
\usepackage[utf8]{inputenc} %
\usepackage[T1]{fontenc}    %
\usepackage{hyperref}       %
\usepackage{url}            %
\usepackage{booktabs}       %
\usepackage{amsfonts}       %
\usepackage{nicefrac}       %
\usepackage{microtype}      %
\usepackage{graphicx}
\usepackage{enumitem}

\usepackage{amssymb}
\usepackage{fdsymbol}
\usepackage{wrapfig}
\usepackage{lipsum}
\usepackage{enumitem}
\usepackage{stackengine}
\usepackage[font=small,labelfont=bf]{caption}
\usepackage{subcaption} % Ensure subfigure or subtable environments are used if this package is included.
\usepackage{color}
\usepackage{adjustbox}

\usepackage{rotating}
\usepackage{makecell}
\usepackage{multirow}
\usepackage{tabularx}
\newcolumntype{Y}{>{\centering\arraybackslash}X}
\newcommand{\cmark}{\textcolor{green!60!black}{\ding{51}}}
\newcommand{\xmark}{\textcolor{red!60!black}{\ding{55}}}

\usepackage{bera}% optional: just to have a nice mono-spaced font
\usepackage{listings}
\usepackage{xcolor}

\definecolor{eclipseStrings}{RGB}{42,0,255}
\definecolor{eclipseKeywords}{RGB}{127,0,85}
\colorlet{numb}{magenta!60!black}
\definecolor{bggray}{rgb}{0.95, 0.95, 0.95}
\definecolor{linenobg}{rgb}{1,1,1}  % white

\lstdefinelanguage{json}{
    basicstyle=\footnotesize\ttfamily,
    commentstyle=\color{eclipseStrings},
    stringstyle=\color{eclipseKeywords},
    numbers=left,
    numberstyle=\scriptsize\color{black}\setlength{\fboxsep}{0pt}\colorbox{linenobg},
    stepnumber=1,
    numbersep=10pt,
    xleftmargin=2em,
    framexleftmargin=2em,
    showstringspaces=false,
    breaklines=true,
    frame=single,
    backgroundcolor=\color{bggray},
    string=[s]{"}{"},
    comment=[l]{:},
    morecomment=[l]{,},
}

\definecolor{dockerbg}{rgb}{0.97, 0.97, 0.97}
\definecolor{scriptbg}{rgb}{0.90, 0.95, 1.00}  % 浅蓝色用于脚本注释行

\PassOptionsToPackage{most}{tcolorbox}
\usepackage[most]{tcolorbox}

\tcbset{
  images/.style={
    colback=dockerbg,                   % 正文背景色
    colframe=gray,           % 整体边框色
    boxrule=1pt,                     % 边框粗细
    arc=0mm,                         % 圆角
    colbacktitle=gray,       % 标题栏背景色
    coltitle=white,                  % 标题文字颜色
    fonttitle=\bfseries\sffamily,    % 标题字体
    title={TITLE},   % 默认标题
    toptitle=2mm,                   % 标题栏内上方留白
    bottomtitle=2mm,                % 标题栏内下方留白
    leftrule=2pt, rightrule=2pt,     % 侧边不额外交框（可按需调整）
    top=1mm, bottom=1mm, left=1mm, right=1mm, % 内边距
  },
  shell/.style={
    colback=blue!5,                   % 正文背景色
    colframe=blue,           % 整体边框色
    boxrule=1pt,                     % 边框粗细
    arc=2mm,                         % 圆角
    colbacktitle=blue,       % 标题栏背景色
    coltitle=white,                  % 标题文字颜色
    fonttitle=\bfseries\sffamily,    % 标题字体
    title={TITLE},   % 默认标题
    toptitle=2mm,                   % 标题栏内上方留白
    bottomtitle=2mm,                % 标题栏内下方留白
    leftrule=2pt, rightrule=2pt,     % 侧边不额外交框（可按需调整）
    top=1mm, bottom=1mm, left=3mm, right=3mm, % 内边距
  }
}

\setlist[itemize]{topsep=0.3em, partopsep=0pt, parsep=0pt, itemsep=0.4em}

\newcommand{\x}{\mathbf{x}}

\definecolor{blanchedalmond}{rgb}{1.0, 0.92, 0.8}
\definecolor{carmine}{rgb}{0.59, 0.0, 0.09}
\definecolor{lightblue}{rgb}{0.22,0.45,0.70}%

\renewcommand{\mathbf}{\boldsymbol}

\makeatletter
\def\Ddots{\mathinner{\mkern1mu\raise\p@
\vbox{\kern7\p@\hbox{.}}\mkern2mu
\raise4\p@\hbox{.}\mkern2mu\raise7\p@\hbox{.}\mkern1mu}}
\makeatother

\definecolor{amaranth}{rgb}{0.9, 0.17, 0.31}
\definecolor{antiquebrass}{rgb}{0.8, 0.58, 0.46}
\definecolor{antiquefuchsia}{rgb}{0.57, 0.36, 0.51}
\definecolor{chromeyellow}{rgb}{0.31, 0.47, 0.26}

\newcommand{\paperlogo}{\raisebox{-1.5pt}{\includegraphics[height=1.3em]{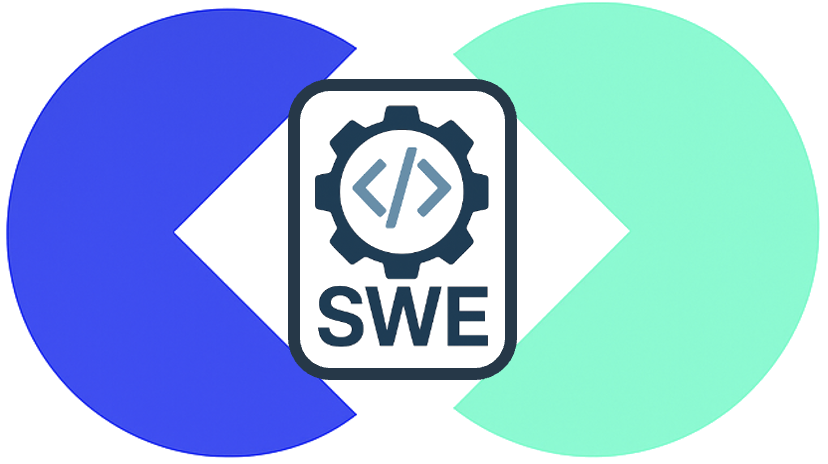}}}

\newtcolorbox{AIbox}[2][]{aibox,title=#2,#1}
\definecolor{lightblue}{rgb}{0.22,0.45,0.70}%
\definecolor{Gray}{gray}{0.95}
\definecolor{Cornsilk}{rgb}{1.0, 0.97, 0.86}

\usepackage{amsmath}

\usepackage[all]{hypcap}
\usepackage{siunitx}
\title{\paperlogo{} Skywork-SWE: Unveiling Data Scaling Laws for Software Engineering in LLMs}

\runningtitle{\paperlogo{} Skywork-SWE: Unveiling Data Scaling Laws for Software Engineering in LLMs}

\author{
  Liang Zeng,
  Yongcong Li,
  Yuzhen Xiao,
  Changshi Li,
  Chris Yuhao Liu,
  Rui Yan,\\
  Tianwen Wei,
  Jujie He,
  Xuchen Song,
  Yang Liu,
  and Yahui Zhou
}
\affil[]{Skywork AI, Kunlun Inc}

\correspondingauthor{Xuchen Song, Yang Liu}

\begin{document}

\begin{abstract}
Software engineering~(SWE) has recently emerged as a crucial testbed for next-generation LLM agents, demanding inherent capabilities in two critical dimensions: sustained iterative problem-solving~(e.g., >50 interaction rounds) and long-context dependency resolution~(e.g., >32k tokens). However, the data curation process in SWE remains notoriously time-consuming, as it heavily relies on manual annotation for code file filtering and the setup of dedicated runtime environments to execute and validate unit tests. Consequently, most existing datasets are limited to only a few thousand GitHub-sourced instances. To this end, we propose an incremental, automated data-curation pipeline that systematically scales both the volume and diversity of SWE datasets. Our dataset comprises \num{10169} real-world Python task instances from \num{2531} distinct GitHub repositories, each accompanied by a task specified in natural language and a dedicated runtime-environment image for automated unit-test validation. We have carefully curated over \num{8000} successfully runtime-validated training trajectories from our proposed SWE dataset. When fine-tuning the Skywork-SWE model on these trajectories, we uncover a striking data scaling phenomenon: \emph{the trained model's performance for software engineering capabilities in LLMs continues to improve as the data size increases, showing no signs of saturation.} Notably, our Skywork-SWE model achieves 38.0\% pass@1 accuracy on the SWE-bench Verified benchmark without using verifiers or multiple rollouts, establishing a new state-of-the-art~(SOTA) among the Qwen2.5-Coder-32B-based LLMs built on the OpenHands agent framework. Furthermore, with the incorporation of test-time scaling techniques, the performance further improves to 47.0\% accuracy, surpassing the previous SOTA results for sub-32B parameter models. Finally, we distill a set of practical guidelines aimed at further advancing LLM-driven software engineering in both academic research and industrial practice.
% We will release our curated dataset, runtime images, training framework, and model checkpoints to accelerate future progress.
We release the Skywork-SWE-32B model checkpoint to accelerate future research.

\vspace{2mm}


\vspace{2mm}

\coloremojicode{1F4C5} \textbf{Date}: June 20, 2025

\coloremojicode{1F3E0} \textbf{Blog}: \href{https://quixotic-sting-239.notion.site/eb17f379610040ceb54da5d5d24065bd}{https://quixotic-sting-239.notion.site/eb17f379610040ceb54da5d5d24065bd}

\coloremojicode{1F917} \textbf{Model Weights}: \href{https://huggingface.co/Skywork/Skywork-SWE-32B}{https://huggingface.co/Skywork/Skywork-SWE-32B}

\coloremojicode{1F4E7} \textbf{Contact}: \href{mailto:liang.zeng@kunlun-inc.com}{liang.zeng@kunlun-inc.com}

\end{abstract}

\maketitle
\vspace{3mm}
\vspace{-4mm}
\section{Introduction}
\label{sec:intro}

\epigraph{\textit{Talk is cheap. Show me the code.}}{--- Linus Torvalds}

\begin{figure*}[!t]
  \centering
  \begin{subfigure}[!t]{0.54\linewidth}
    \centering
    \includegraphics[width=\linewidth]{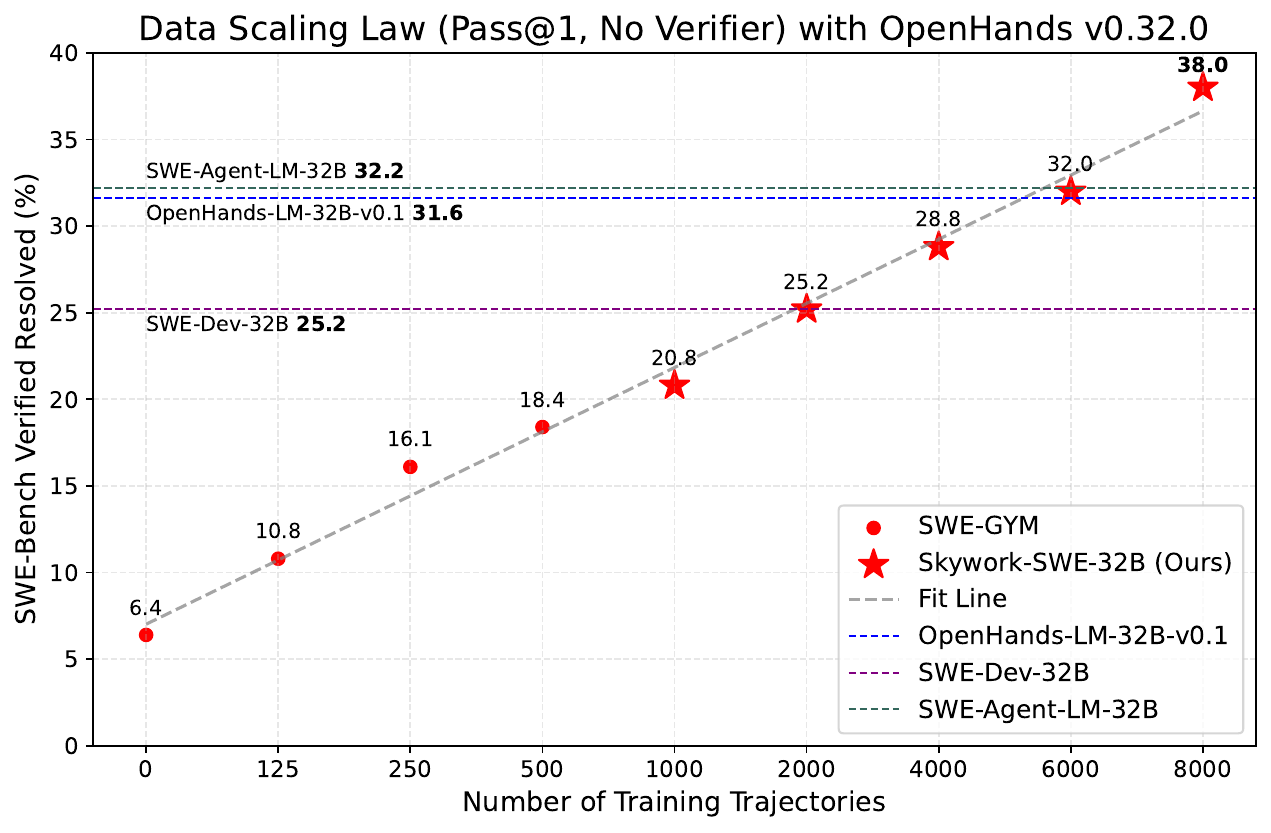}
    \label{fig:data_scaling_a}
    \vspace{-1em}   
  \end{subfigure}
   
  \begin{subfigure}[!t]{0.89\linewidth}
    \centering
    \includegraphics[width=0.61\linewidth]{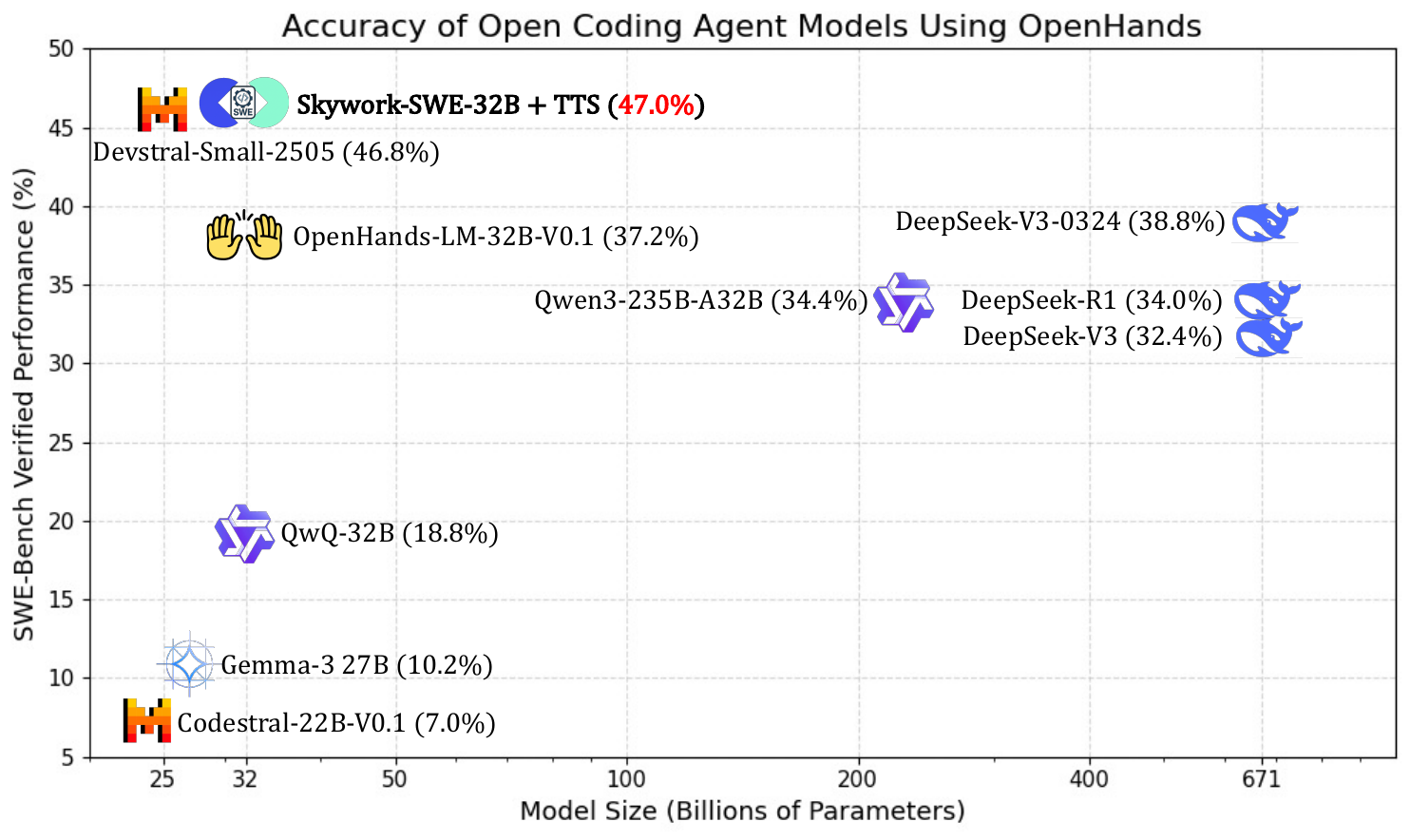}
    \label{fig:data_scaling_b}
  \end{subfigure}
  \caption{\textbf{(Top)} The Skywork-SWE model achieves 38.0\% pass@1 accuracy on the SWE-bench Verified benchmark, outperforming previous open-source SoTA Qwen2.5-Coder-32B-based LLMs built on the OpenHands agent framework. Moreover, Skywork-SWE clearly demonstrates the data scaling law phenomenon for software engineering capabilities in LLMs, with no signs of saturation at \num{8209} collected training trajectories. \emph{All evaluations are conducted in-house using a single attempt per instance, without verifiers or multiple rollouts, based on the OpenHands v0.32.0 framework.}
  \textbf{(Bottom)} Performance comparison among recent advanced approaches using OpenHands on SWE-bench Verified. With the incorporation of test-time scaling~(TTS) techniques, Skywork-SWE achieves a further improvement to 47.0\% accuracy, surpassing the latest Qwen and DeepSeek model series.
  }
  \vspace{-1em} 
  \label{fig:data_scaling}
\end{figure*}
Two core capabilities define the emerging potential of Large Language Model~(LLM) agents: the ability to engage in multi-turn interactions and to reason over long-context inputs~\citep{openai2025o3o4card, team2023gemini, guo2025deepseek, weng2023agent}. Among real-world applications, software engineering~(SWE) tasks~\citep{jimenez2024swebench}, which involve localizing bugs, modifying source codes, and validating fixes on real-world software issues collected from GitHub, stand out as a critical evaluation domain. Unlike conventional code generation tasks, which produce simple code snippets to solve competitive programming problems~\citep{jain2024livecodebench,zhuo2024bigcodebench}, SWE tasks require iterative problem-solving over extended interaction rounds to utilize code tools and the ability to manage long-context dependencies within code files to address the real-world software engineering challenges~\citep{pan2024training,yang2025swe}. 
The increasing prominence of benchmarking datasets such as SWE-bench~\citep{jimenez2024swebench}, SWE-bench Verified~\citep{openai2024swebench} reflects both the rising research interest and the inherent challenges of LLM-driven SWE.

\begin{table}[t]
  \centering
  \caption{Comparison of publicly available benchmark datasets for SWE tasks. \textit{Executable Environment}: whether each instance is provided with an executable environment in which all necessary dependencies are pre-installed. \textit{Verified Unit Tests}: whether the associated unit tests have been validated to ensure the correctness of both \texttt{FAIL\_TO\_PASS} and \texttt{PASS\_TO\_PASS} outcomes. \textit{Code Execution Suite}: whether a unified and automated script is provided to automatically initialize, configure, and execute tests across diverse repositories without manual intervention. The Skywork-SWE dataset encompasses all three critical dimensions of SWE benchmarks and is systematically scaled to highlight both the volume~(\#Instances) and diversity~(\#Repos) of SWE datasets.}
  \footnotesize
  \setlength{\tabcolsep}{6pt}
  \begin{adjustbox}{max width=\linewidth}
  \begin{tabular}{l|ccc|rr}
    \toprule
    Dataset & Executable Env. & Verified Unit Tests & Code Exec. Suite & \#Instances & \#Repos \\
    \midrule
    SWE-bench \citep{jimenez2024swebench} & \cmark & \cmark & \xmark & \num{2294} & 12 \\
    SWE-bench Lite \citep{jimenez2024swebench} & \cmark & \cmark & \xmark & 300 & 12 \\
    SWE-bench Verified \citep{openai2024swebench} & \cmark & \cmark & \xmark & 500 & 12 \\
    \midrule
    SWE-bench Extra \citep{badertdinov2024scaling} & \xmark & \xmark & \cmark & \num{6376} & \num{1974} \\
    SWE-Fixer \citep{xie2025swe} & \xmark & \xmark & \xmark & \num{115406} & 856 \\
    SWE-Smith \citep{yang2025swe} & \cmark & \cmark & \xmark & \num{50137} & 128 \\
    SWE-Gym \citep{pan2024training} & \cmark & \cmark & \xmark & \num{2438} & 11 \\
    \midrule
    \textbf{Skywork-SWE} & \cmark & \cmark & \cmark & \num{10169} & \num{2531} \\
    \bottomrule
  \end{tabular}
  \end{adjustbox}
  \label{tab:stats_diff_dataset}
  \vspace{-1em}
\end{table}

Despite these advances, existing datasets still suffer from key limitations that hinder progress in the field:
\begin{itemize}[leftmargin=*]
    \item \textbf{Insufficient environment and validation support.} As detailed in Table~\ref{tab:stats_diff_dataset}, existing benchmarks typically lack comprehensive mechanisms for configuring executable runtime environments or standardized code execution suites to systematically validate the generated code patches across diverse repositories. For example, SWE-bench-extra~\citep{badertdinov2024scaling} and SWE-Fixer~\citep{xie2025swe} either omit executable environments entirely or lack rigorous test validations, resulting in inconsistent and non-reproducible evaluations.
    \item \textbf{Scarcity of high-quality training data.} Although some existing datasets are large in scale, few provide rigorously validated and high-quality training instances. This lack of openly available validated data has led open-source LLMs to consistently underperform compared to proprietary models on SWE tasks. For example, SWE-Dev~\citep{wang2025swedev} lacks rigorously validated training instances, and SWE-Gym~\citep{pan2024training} suffer from limited repository coverage.
    \item \textbf{Unclear applicability of data scaling laws.} The volume of training data for SWE tasks is notably smaller than that in other LLM domains~\citep{pan2024training}. It remains uncertain whether data scaling laws~\citep{Kaplan2020,Hoffmann2022} still hold in the software engineering context. Addressing this question is crucial for guiding future dataset expansion and optimizing model training strategies.
\end{itemize}

In response to these challenges, we develop an automated data curation pipeline that systematically scales both the volume and diversity of Skywork-SWE datasets. As shown in Table~\ref{tab:stats_diff_dataset}, the Skywork-SWE dataset comprises \num{10169} real-world Python task instances drawn from \num{2531} GitHub repositories. Each instance in the Skywork-SWE dataset includes a detailed natural language description along with an associated executable runtime environment specifically designed to support automated execution and test validation. We propose a three-stage incremental pipeline consisting of data collection and pre-filtering, execution-based validation, and agent trajectory generation for SWE tasks within a unified framework. This approach ensures the high quality and diversity of the Skywork-SWE dataset by combining broad coverage of GitHub repositories with rigorous reproducibility.

Leveraging this dataset, we fine-tune our Skywork-SWE model on over \num{8000} successfully validated trajectories, achieving 38.0\% pass@1 accuracy on the SWE-bench Verified benchmark \textit{without using verifiers or multiple rollouts}. This establishes a new state-of-the-art among the Qwen2.5-Coder-32B-based LLMs built on the OpenHands agent framework. As shown in Fig.~\ref{fig:data_scaling}, our extensive experiments reveal a clear data-scaling trend: the trained model's software engineering capabilities consistently improve with increased training data size, validating the applicability of scaling laws to SWE tasks. Moreover, applying test-time scaling techniques boosts performance to 47.0\% accuracy, outperforming prior SoTA results for LLMs with fewer than 32 billion parameters.

Skywork-SWE aims to bridge the gap between open-source and proprietary SWE agent models, fostering transparency, reproducibility, and progress in LLM-driven software engineering. Our main contributions can be summarized as follows:
\begin{itemize}[leftmargin=*]
    \item We propose an efficient and automated pipeline for SWE data collection, resulting in the Skywork-SWE dataset, a large-scale, high-quality dataset featuring comprehensive executable runtime environments.
    \item We release Skywork-SWE-32B, a powerful open-source code agent model tailored for SWE tasks, establishing a new performance benchmark among the same-scale open-source SWE agents.
    \item We empirically observe the data scaling law in SWE tasks, demonstrating consistent performance improvements with increased training data size. This finding not only validates the applicability of scaling laws within software engineering but also highlights the need for larger, well-curated datasets to further enhance model performance.
\end{itemize}

\section{Related Work}
\label{sec:related_work}

\subsection{Code-related Tasks in LLMs}
Large Language Models~(LLMs) have achieved remarkable progress across a wide range of code-related tasks~\citep{jiang2024survey, chang2024systematic}, ranging from code snippet generation~\citep{leblond2023alphacode2} to complex software engineering tasks~\citep{jimenez2024swebench}. Below, we briefly outline these two major directions of LLMs in code-related tasks.

\textbf{Code Generation} empowers LLMs to synthesize functional programs from natural language descriptions. Early function-level benchmarks such as HumanEval~\citep{chen2021evaluating} and MBPP~\citep{austin2021program} laid the groundwork for this task, spurring the development of code-oriented LLMs including AlphaCode~\citep{leblond2023alphacode2}, Code Llama~\citep{roziere2023code}, WizardCoder~\citep{luo2023wizardcoder}, StarCoder~\citep{li2023starcoder}, and DeepSeek-Coder~\citep{deepseek2024coder}. Through curated training datasets and specialized fine-tuning techniques, these models exhibit strong performance on code generation tasks and have nearly saturated the performance limits of traditional benchmarks. To advance evaluation, recent benchmarks such as LiveCodeBench~\citep{jain2024livecodebench} and BigCodeBench~\citep{zhuo2024bigcodebench} introduce contamination-free, practical, and challenging programming problems, offering a more rigorous assessment of the code capabilities of modern LLMs~\citep{jaech2024openai, openai2024reasoning}.
In response, leading large reasoning models (LRMs), such as OpenAI’s o3~\citep{openai2025o3o4card}, DeepSeek-R1~\citep{guo2025deepseek} and Kimi-k1.5~\citep{team2025kimi}, employ reinforcement learning to incentive chain-of-thought reasoning capability of LLMs to substantially improve performance on these code benchmarks.

\textbf{Software Engineering}~(SWE) focuses on resolving real-world GitHub issues  within repository environments, where agents are required to localize bugs, modify source code, and validate fixes based on execution results. This practical task marks a shift in the field---from static, single-turn code generation to dynamic, interactive coding workflows---significantly expanding the applicability and capabilities of LLMs in real-world software development scenarios.
SWE-bench~\citep{jimenez2024swebench} and its successor, SWE-bench Verified~\citep{openai2024swebench}, serve as the canonical benchmark in this domain. They provide thousands of real GitHub issues accompanied by complete codebases, natural language descriptions, and human-filtered regression test suites to reliably evaluate LLM models’ ability to solve real-world software issues.
Creating training data for SWE agents is a difficult and labor-intensive process. Recently, numerous efforts have been devoted to synthesizing training data for SWE tasks, with each instance required to be validated in its corresponding runtime environment.
SWE-Gym~\citep{pan2024training} provide executable environments featuring realistic SWE tasks and corresponding unit tests, albeit with a relatively limited scale of just over \num{2000} instances. SWE-bench-extra~\citep{badertdinov2024scaling} extends the methodology used in constructing the SWE-bench benchmark~\citep{jimenez2024swebench}, resulting in \num{6415} Python issue-pull request instances. SWE-Dev~\citep{wang2025swedev} proposes a test-case construction pipeline using structured descriptions and execution validation to bypass the overhead associated with full runtime verification. 
To enable LLM-based agents to learn from large-scale SWE data, SWE-fixer~\citep{xie2025swe} and SWE-Smith~\citep{yang2025swe} synthetically generate \texttt{Fail\_to\_Pass} instances by injecting and validating plausible bugs, resulting in thousands of SWE task instances. 
Furthermore, reinforcement learning approaches---as exemplified by SkyRL~\citep{cao2025skyrl}---combine asynchronous rollouts with verifier-based feedback, promoting long-horizon resolution strategies in SWE tasks. 

In this work, we propose new SWE data corpus consisting of over \num{10000} real-world Python task instances collected from \num{2531} distinct GitHub
repositories. Each instance is paired with a dedicated runtime environment image to enable automated unit test validation. Leveraging this large-scale dataset, our Skywork-SWE model demonstrates clear evidence of the data scaling law in software engineering tasks.

\subsection{Code Agent Framework in Software Engineering}
Recent advancements in \textbf{software engineering (SWE) agents} have moved beyond generic “toy” code generation toward \emph{repository-scale debugging and build orchestration}.
OpenHands~\citep{wang2024openhands} provides an open, event-driven platform that enables LLM agents iteratively to edit files, execute shell commands and browse the Web inside sandboxed containers, establishing strong and reproducible baselines on SWE-bench. 
In contrast to interactive multi-step planners, Agentless~\citep{xia2024agentless} demonstrates that a streamlined \emph{localise–repair–validate} pipeline can outperform many SWE agents on the SWE-bench benchmark while reducing costs by an order of magnitude.
Moatless~\citep{Orwall2024Moatless} contends that effective context retrieval, rather than complex reasoning loops, is the key to patch generation; its open-source toolkit showcases scalable prompting and edit application across million-line codebases.
Focusing on learning-based approaches, SWE-Fixer~\citep{xie2025swe} integrates coarse-to-fine file retrieval and edit disentanglement to train open LLMs for SWE tasks.
To enable LLM agents to autonomously operate computer interfaces for solving SWE tasks,  \textit{SWE-Agent}~\citep{yang2024swe} further formalises an \emph{Agent-Computer Interface} (ACI) that exposes editor, shell and test runners as structured actions.
Collectively, these frameworks highlight three emerging trends: (i)~\emph{lean pipelines} that replace heavy planning with specialized tools (Agentless, Moatless); (ii)~\emph{retrieval-aware fine-tuning} for efficient patch synthesis (SWE-Fixer); and (iii)~\emph{rich ACIs} that enable full-stack repository manipulation (OpenHands, SWE-Agent). 

In this study, we adopt OpenHands~\citep{wang2024openhands} as our code agent framework due to its strong empirical performance and widespread adoption in the open-source code agent community. Building upon this foundation, our Skywork-SWE model achieves the SOTA performance on SWE-bench Verified with test-time scaling techniques for sub-32B parameter models in the open-sourced code agent ecosystem.

\section{Method}
\label{sec:method}

\begin{figure*}[t]
    \centering\includegraphics[width=\linewidth]{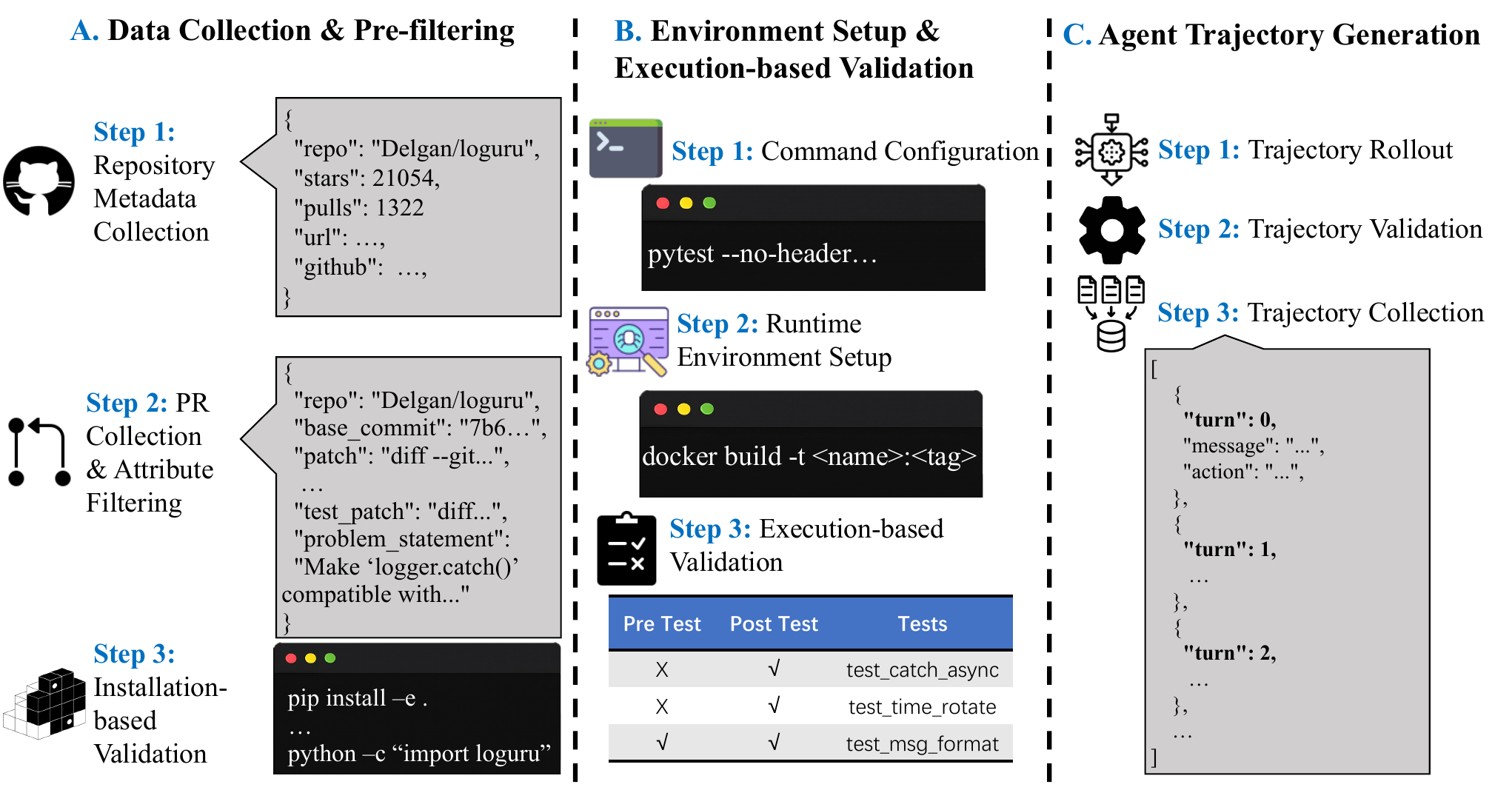}
    \caption{Overview of our three-stage Skywork-SWE data collection pipeline. (1) \textbf{Stage A.} \textbf{Data Collection and Pre-filtering}: Step A.1 scrapes GitHub repo metadata. Step A.2 collects and filters relevant pull requests~(PRs). Step A.3 validates PRs via installation checks. (2) \textbf{Stage B.} \textbf{Execution-based Validation}: Step B.1 configures unified execution commands. Step B.2 builds Docker-based runtime environments. Step B.3 validates task instances by executing unit tests. (3) \textbf{Stage C.} \textbf{Agent Trajectory Generation}: Step C.1 generates agent trajectories. Step C.2 validates trajectories via patch testing. Step C.3 collects validated trajectories for training.}
    \label{fig:pipeline}
\end{figure*}

In this section, we present the three-stage Skywork-SWE data collection pipeline used to construct our high-quality Skywork-SWE dataset. As shown in Fig.~\ref{fig:pipeline}, the pipeline consists of: (1) \textbf{data collection and pre-filtering} (Sec.~\ref{sec:data}), (2) \textbf{environment setup and execution-based validation} (Sec.~\ref{sec:3.2}), and (3) \textbf{agent trajectory generation} (Sec.~\ref{sec:trajectory}). We also provide a flow diagram in Fig.~\ref{fig:data_flow} to illustrate how data volume evolves throughout the pipeline. The curated data is subsequently used to train our Skywork-SWE-32B agent model, enhancing its ability to tackle software engineering tasks (Sec.~\ref{sec:train}).  

\subsection{Data Collection and Pre-filtering}
\label{sec:data}
\begin{figure*}[t]
    \centering
    \includegraphics[width=\linewidth]{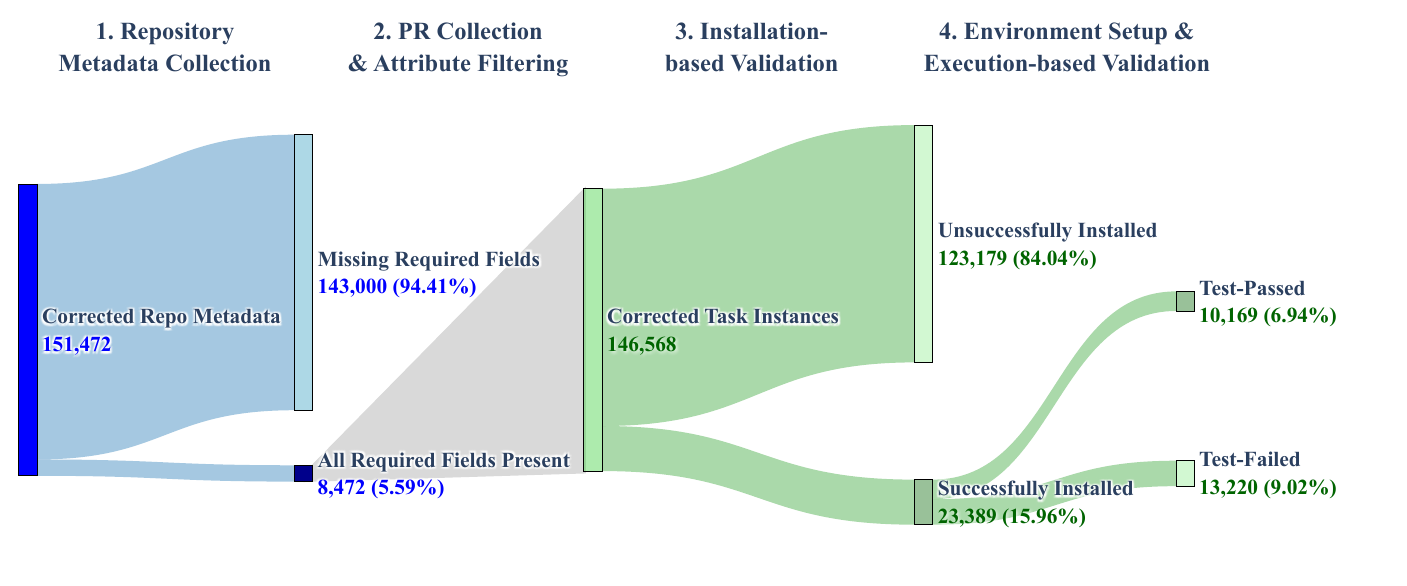}
    \vspace{-3em}
    \caption{Visualization of data flow across four key hierarchical filtering steps in our data collection pipeline. The first three steps belong to the Data Collection \& Pre-filtering stage in Fig.~\ref{fig:pipeline}, while the last step corresponds to the Environment Setup \& Execution-based Validation stage. The text in \textcolor{blue}{Blue} and \textcolor{green}{green} indicates the number of repositories and task instances, respectively.}
    \label{fig:data_flow}
    \vspace{-0.6em}
\end{figure*}
The volume of high-quality training data for SWE tasks remains remarkably smaller than that in other domains~\citep{wang2025swedev}. Previous work has shown that high-quality and diverse training data are crucial for efficiently training code agent models~\citep{pan2024training, kaplan2020scaling}. In this stage, we collect a large number of task instances from GitHub repositories and perform an initial round of data filtering.

\textbf{Repository metadata collection.} We collect metadata from a large number of open repositories on GitHub using the GitHub developer API. \emph{To avoid data leakage when collecting GitHub repositories, we exclude those repositories already included in SWE-Bench Verified.} As shown in Fig. \ref{fig:pipeline}, the key fields in the metadata include \texttt{repo}, which denotes the repository's \texttt{owner/name} identifier, and \texttt{stars}, which represent the number of stars the specific code repository has received. We prioritize GitHub repositories with higher \texttt{stars} counts. As depicted in Fig. \ref{fig:data_flow}, we scrape metadata for a total of \num{151472} repositories. After filtering out the entries with missing required fields, we retain \num{8472} valid metadata entries.

\textbf{Pull request collection and attribute filtering.} Using the metadata collected in the previous step, we scrape pull request (PR) data from the corresponding repositories via the GitHub developer API to form the initial task instances. Following the practice of SWE-bench \citep{jimenez2024swebench}, we select only merged PRs that link to and resolve a GitHub issue using keywords such as ``closes/fixes/resolves \#…''. We retain only those PRs that modify test-related files (i.e., any path or filename containing ``test'' or ``testing''). As shown in Fig. \ref{fig:pipeline}, the key fields for these task instances include \texttt{base\_commit} (the commit hash of the repository representing the HEAD before the solution PR is applied), \texttt{patch} (the gold patch generated by the PR that resolved the issue), and \texttt{test\_patch} (a test-file patch included in the solution PR). As shown in Fig. \ref{fig:data_flow}, we scrape a total of \num{146568} initial task instances.

\textbf{Installation-based validation.} We filter the task instances by reverting the repositories to their \texttt{base\_commit} and executing predefined installation commands in a base environment. As shown in Fig.~\ref{fig:data_flow}, we discard \num{123179} instances that fail to install, resulting in \num{23389} valid instances.

\subsection{Execution-based Validation and Runtime Environment Setup}
\label{sec:3.2}

The execution-based validation of candidate task instances relies on constructing dedicated runtime environments that satisfy the dependencies of a wide range of code libraries. Due to significant heterogeneity across repositories, manual configuration is time-consuming and difficult to scale. At this stage, we implement an automated per-instance pipeline to streamline the process.

\subsubsection{Command Configuration} 

\begin{figure}[!t]
\centering
\begin{lstlisting}[language=json,firstnumber=1]
{
	"python": "3.9",
	"packages": "requirements.txt",
	"pip_packages": ["pytest", "hypothesis", "mock", "setuptools", ...],
	"install": "pip install -e . || true; pip install -e .[test] ...",
	"pre_install": "apt update && apt install -y make gcc g++ pkg-config",
\end{lstlisting}
\vspace{-1em}
\begin{lstlisting}[language=json,firstnumber=7]
	"test_cmd": "pytest --no-header -rA --tb=no ..."
}
\end{lstlisting}
\caption{The default configuration snippet specifying environment setup, dependency installation~(lines 2-6), and test execution commands~(line 7) for all task instances.}
\label{list:default_configure}
\vspace{-0.5em}
\end{figure}

To standardize execution-based validation and runtime environment setup across diverse repositories, we define a unified default setup that includes environment initialization, dependency installation, and test command execution. Traditional approaches, such as that used in SWE-bench~\citep{jimenez2024swebench}, require manually specifying separate configuration commands for each repository. However, due to the substantial variability in test environments across repositories and commits, this manual process is highly resource-intensive and difficult to scale. To address this, we adopt a generalizable configuration strategy inspired by SWE-bench-extra~\citep{badertdinov2024scaling}, designed to maximize compatibility across a broad range of pull-request instances in various GitHub repositories.

The configuration comprises several components, as illustrated in Fig.~\ref{list:default_configure}. It sets Python~3.9 as the default runtime and installs essential system packages (e.g., \texttt{make}, \texttt{gcc}, \texttt{g++}, \texttt{pkg-config}) required for building native extensions. 
Python dependencies are installed via the \texttt{requirements.txt} file, supplemented by commonly used development and testing packages such as \texttt{pytest}, \texttt{hypothesis}, \texttt{mock}, and \texttt{setuptools}. To accommodate varying naming conventions across GitHub repositories, the configuration also includes fallback installation commands for optional extras like \texttt{[test]}, \texttt{[tests]}, and \texttt{[dev]}. Test execution is standardized using a unified \texttt{pytest} command that disables cache usage and suppresses deprecation warnings to produce consistent and reproducible results.

\subsubsection{Runtime Environment Setup} 

\begin{figure}[!t]
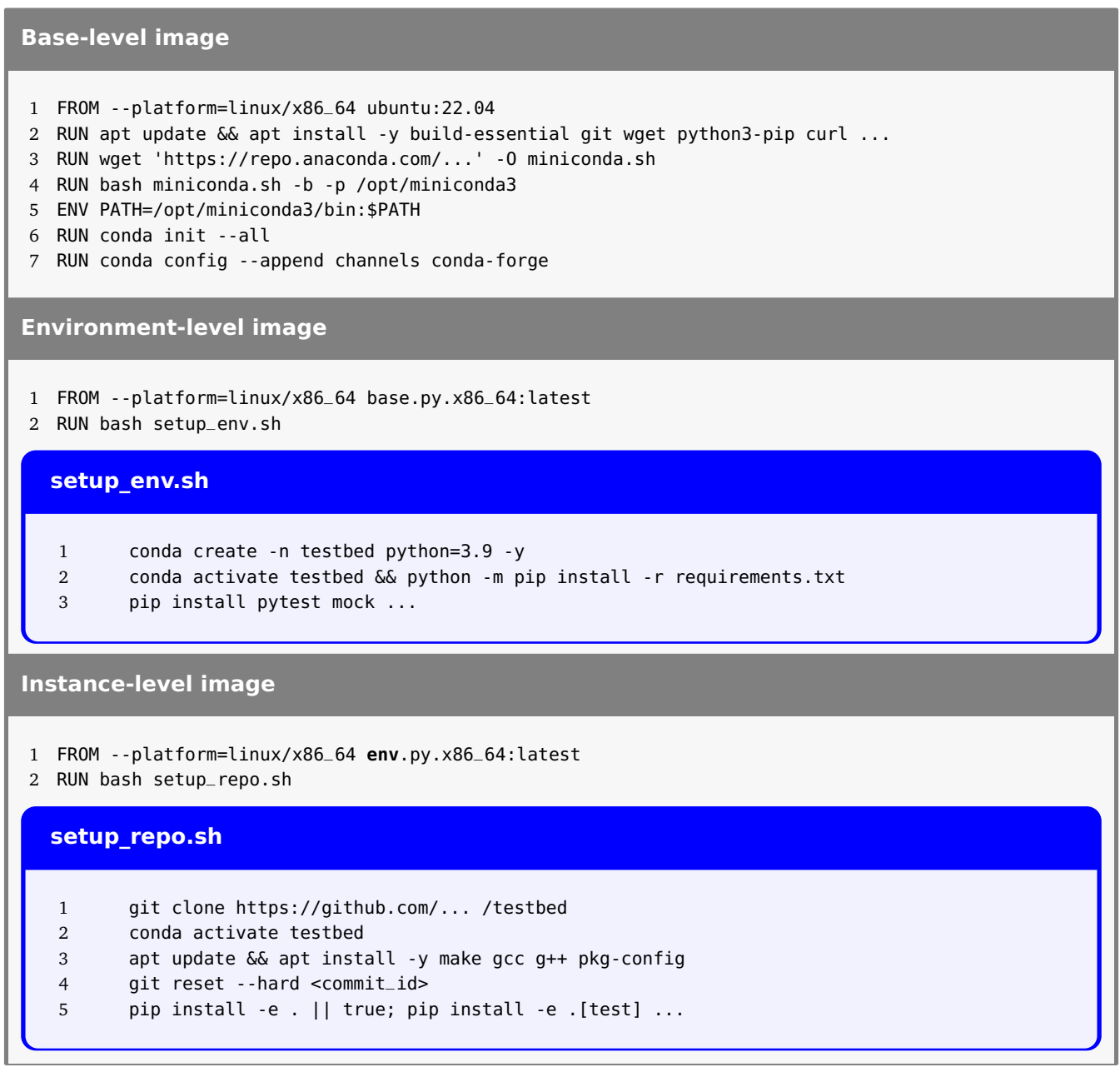

\centering
\begin{tcolorbox}[images, title={Base-level image}]
  \begin{lstlisting}[language=bash]
FROM --platform=linux/x86_64 ubuntu:22.04
RUN apt update && apt install -y build-essential git wget python3-pip curl ...
RUN wget 'https://repo.anaconda.com/...' -O miniconda.sh
RUN bash miniconda.sh -b -p /opt/miniconda3
ENV PATH=/opt/miniconda3/bin:$PATH
RUN conda init --all
RUN conda config --append channels conda-forge
  \end{lstlisting}
\end{tcolorbox}
  \vspace{-1.5em}
\begin{tcolorbox}[images, title={Environment-level image}]
  \begin{lstlisting}[language=bash]
FROM --platform=linux/x86_64 base.py.x86_64:latest
RUN bash setup_env.sh
  \end{lstlisting}
    \begin{tcolorbox}[shell, title={setup\_env.sh}]
     \begin{lstlisting}[language=bash, frame=none, keywordstyle=\normalfont\ttfamily, backgroundcolor=\color{blue!5}]
    conda create -n testbed python=3.9 -y
    conda activate testbed && python -m pip install -r requirements.txt
    pip install pytest mock ...
        \end{lstlisting}
    \end{tcolorbox}
\end{tcolorbox}
  \vspace{-1.5em}
\begin{tcolorbox}[images, title={Instance-level image}]
  \begin{lstlisting}[language=bash, label={fig:docker-three-layer}]
FROM --platform=linux/x86_64 env.py.x86_64:latest
RUN bash setup_repo.sh
  \end{lstlisting}
    \begin{tcolorbox}[shell, title={setup\_repo.sh}]
    \begin{lstlisting}[language=bash, frame=none, keywordstyle=\normalfont\ttfamily, backgroundcolor=\color{blue!5}]
    git clone https://github.com/... /testbed
    conda activate testbed
    apt update && apt install -y make gcc g++ pkg-config
    git reset --hard <commit_id>
    pip install -e . || true; pip install -e .[test] ...
    \end{lstlisting}
    
    \end{tcolorbox}
\end{tcolorbox}
\caption{Dockerfile snippets illustrating the construction of a three-level Docker image structure, comprising a base-level image, an environment-level image, and an instance-level image.}
\label{list:image_layers}
\vspace{-1em}
\end{figure}

We automate the construction of execution environments for candidate task instances by using Docker to build isolated runtime images based on a unified configuration defined in the command configuration step. To minimize redundant computation, images are incrementally built to enable reuse of base images for each instance. Meanwhile, successfully built images are cached to accelerate subsequent validations. As illustrated in Fig.~\ref{list:default_configure}, the image-building process is governed by a default configuration file that specifies core components such as the Python version, system packages, and dependency installation commands. The resulting Docker image follows a three-level hierarchy---base-level, environment-level, and instance-level images---each corresponding to a distinct build stage. Simplified versions of the scripts used for the three-level Docker image construction are listed in Fig.~\ref{list:image_layers}.
\begin{itemize}[leftmargin=*]
    \item The base-level image is built from Ubuntu 22.04 with the \texttt{linux/x86\_64} platform specified for compatibility. It installs essential system packages (e.g., \texttt{build-essential}, \texttt{git}, \texttt{wget}) via \texttt{apt}, followed by the installation of Miniconda and configuration of the \texttt{conda-forge} channel for Python package management.
    \item The environment-level image builds upon the base image by executing the \texttt{setup\_env.sh} script, which creates a Conda environment with Python~3.9 (as specified in the \texttt{"python"} field), installs dependencies from the \texttt{requirements.txt}, adds additional development and testing packages listed under \texttt{"pip\_packages"}.
    \item The instance-level image further extends the environment by executing \texttt{setup\_repo.sh}, which clones the target repository, checks out a specific commit, installs system dependencies listed under the \texttt{"pre\_install"} field, and performs an editable installation of the corresponding repository with optional extras defined in the \texttt{"install"} key.
\end{itemize}

\subsubsection{Execution-based Validation} 
We evaluate each candidate task instance by applying its patches and executing the repository’s test suite using the corresponding instance-level Docker image constructed during the runtime environment setup step. The validation process consists of two steps:
\begin{itemize}[leftmargin=*]
    \item \textbf{Empty test}: Apply the \texttt{test\_patch} to the \texttt{base\_commit} and run the test suite. Failing tests are labeled as \texttt{empty-FAIL}, and passing ones as \texttt{empty-PASS}.
    \item \textbf{Gold test}: Apply both the \texttt{test\_patch} and the generated code \texttt{patch}, then rerun the test suite. Failing tests are labeled as \texttt{gold-FAIL}, and passing ones as \texttt{gold-PASS}.
\end{itemize}

We define \texttt{FAIL\_TO\_PASS} as the set of tests that fail in the empty test (\texttt{empty-FAIL}) but pass in the gold test (\texttt{gold-PASS}), and \texttt{PASS\_TO\_PASS} as those passing in both stages. Only instances with a non-empty \texttt{FAIL\_TO\_PASS} set are retained, indicating that the applied patch resolves at least one failing test case and successfully corrects previously failing behavior. As shown in Fig.~\ref{fig:data_flow}, out of \num{23389} candidates, \num{13220} instances are filtered out during validation, resulting in a final dataset of \num{10169} verified instances.

\subsubsection{Dataset Statistics}

\begin{figure}[ht]
  \centering
  \begin{minipage}[t]{0.485\textwidth}
    \vspace{0pt}
    \footnotesize
    \setlength{\tabcolsep}{4pt}
    \begin{tabularx}{\textwidth}{l l Y Y Y}
      \toprule
      {Category} &{Metric} &
      {Skywork-SWE} &
      {SWE-Gym Lite} &
      {SWE-bench Verified} \\
    \midrule
      \multirow{2}{*}{Size}
                 & \# Instances      & {\num{10169}}     & 230   & 500     \\
                 & \# Repos          &  \num{2531}     &   11   &    12     \\
      \midrule
      Issue Text & \# Words   &   140.3    &  186.2 & 189.3    \\
      \midrule
      \multirow{2}{*}{Hints Text}
       & \# Words   &  62.2    & 151.4 &  151.9    \\
       & \# Hints   &  \num{2459}    & 155 & 338   \\
      \midrule
      \multirow{4}{*}{Gold Patch}
                 & \# Files edited   &     2.5    &    1.0  &    1.2    \\
                 & \# Func.\ edited  &     2.3    &    1.4  &    2.1    \\
                 & \# Hunks edited  &     6.0    &    1.6  &    2.4    \\
                 & \# Lines edited   &    74.2    &   10.6  &   14.3   \\
      \midrule
      \multirow{3}{*}{Tests}
                 & \# Fail to Pass   &    10.2    &  2.0  &    3.0    \\
                 & \# Pass to Pass   &    86.2    &  99.9  &  120.3    \\
                 & \# Total          &    96.4    &  101.9  &  123.3    \\
      \bottomrule
    \end{tabularx}
\captionof{table}{Dataset statistics for Skywork-SWE, SWE-Gym Lite~\citep{pan2024training}, and SWE-bench Verified~\citep{openai2024swebench}. 
All metrics are reported as per-instance averages, except for those under the Size category and the \#Hints metric.}
    \label{tab:stats}
  \end{minipage}
  \hfill
  \begin{minipage}[t]{0.48\textwidth}
    \vspace{0pt}
    \centering
    \includegraphics[width=\linewidth]{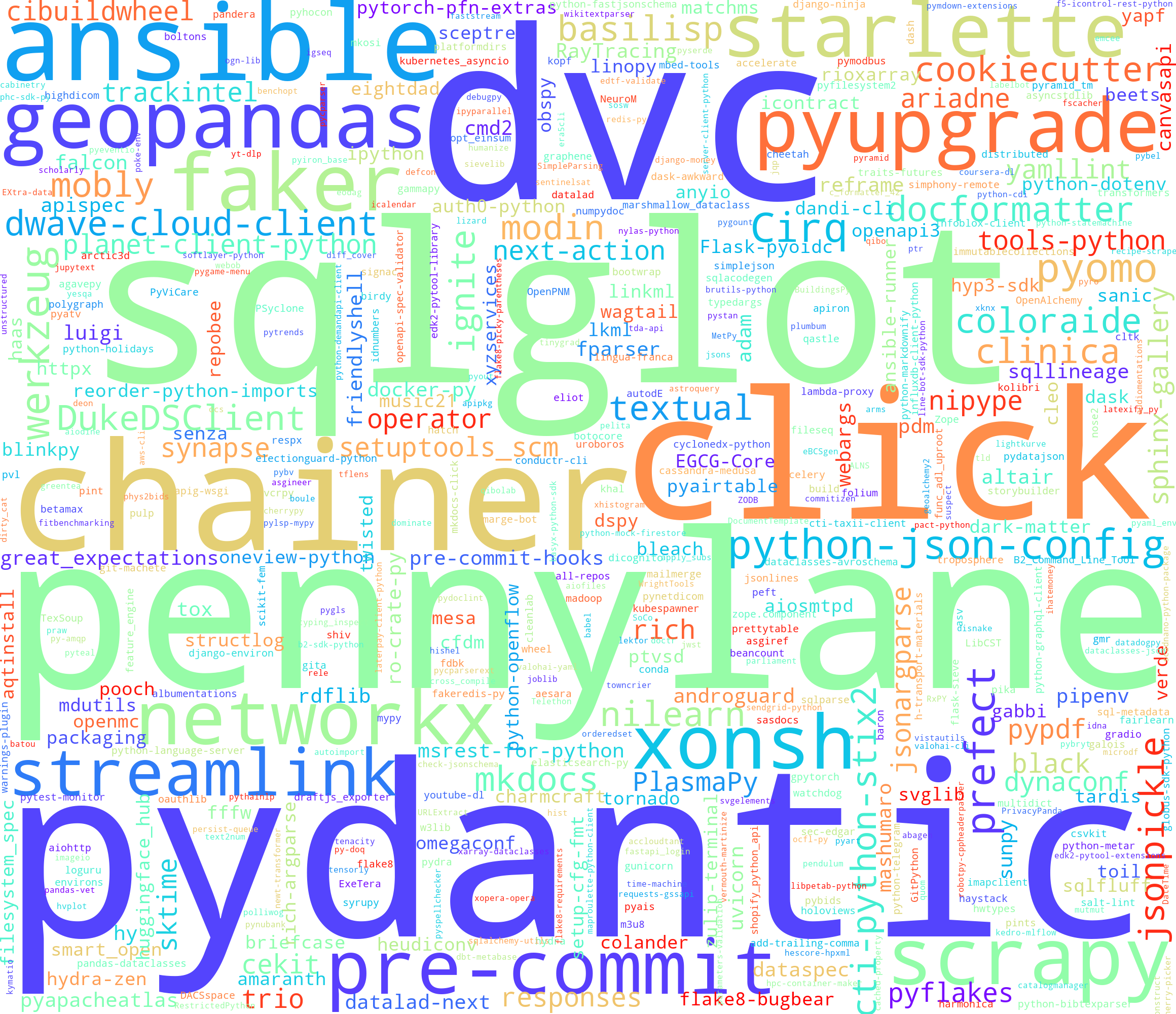}
    \captionof{figure}{Word cloud of repository names in the Skywork-SWE dataset. Font size is proportional to the number of instances in each repository. The Skywork-SWE dataset demonstrates high diversity across the collected GitHub repositories.}
    \label{fig:wordcloud}
  \end{minipage}
\end{figure}

The final Skywork-SWE dataset comprises \num{10169} validated instances collected from \num{2531} unique repositories, obtained through the execution-based validation process. Each instance is paired with a dedicated Docker image to support reproducible execution. On average, each image occupies approximately \num{1.2}~GB, resulting in a total storage footprint of around \num{11.9}~TB. We summarize key statistics and characteristics of the Skywork-SWE dataset as below.

\textbf{Scale.} Skywork-SWE significantly exceeds the scale of existing benchmarks as shown in Table~\ref{tab:stats}. It includes \num{10169} validated instances—more than 20 times the size of SWE-Gym Lite and SWE-bench Verified, which contain 230 and 500 instances, respectively. These instances span over \num{2500} repositories, whereas previous datasets include at most 12 repositories. This broader coverage introduces a wider range of real-world software engineering scenarios.

\begin{figure}[t]
    \centering
    \includegraphics[width=0.95\linewidth]{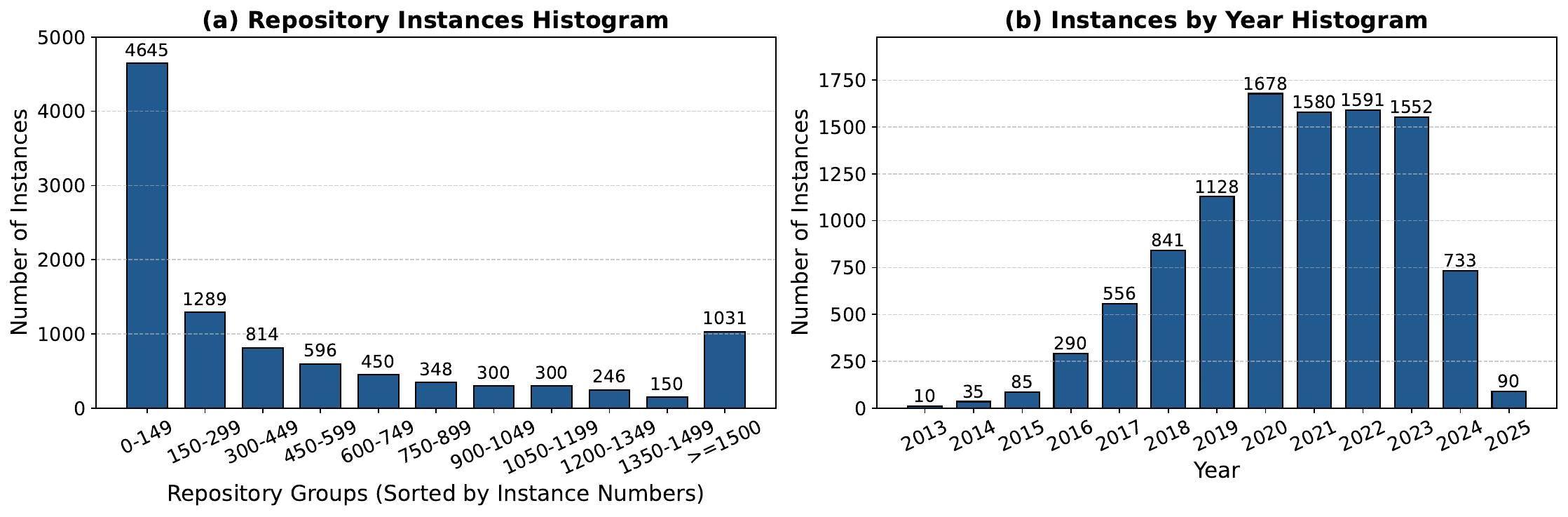}
    \vspace{-3mm}
    \caption{Repository- and year-wise histograms on the Skywork-SWE dataset. (a) The x-axis denotes the number of instances per repository, with every 150 repositories grouped into a single bin for clarity. (b) The x-axis indicates the year in which each issue was created, reflecting the temporal distribution of instance counts. The Skywork-SWE dataset reflects substantial diversity in both temporal coverage and repository sources.}
    \vspace{-0.5em}
    \label{fig:repo_year_analysis}
\end{figure}

\textbf{Diversity.} The dataset exhibits substantial diversity in GitHub repository sources and follows a long-tailed distribution.  Fig.\ref{fig:wordcloud} reflects this diversity, showcasing both well-known projects (e.g., \textit{pydantic}, \textit{dvc}, \textit{sqlglot}, \textit{pennylane}) and a wide array of smaller repositories, indicating a broad representation of software projects. Fig.~\ref{fig:repo_year_analysis}(a) further reveals a pronounced long‐tailed distribution of instance counts across repositories. Approximately 450 repositories (roughly 4.4\%) account for over 66\% of all instances, whereas the remaining \num{9719} repositories contribute less than 34\%. Notably, more than \num{9000} repositories contain fewer than three instances each, significantly enhancing the dataset’s diversity.

\textbf{Temporal Distribution.} The histogram in Fig.~\ref{fig:repo_year_analysis}(b) illustrates the annual distribution of dataset instances from 2013 to 2025. In the early years~(2013–2015), the number of instances remains relatively low~(10, 35, and 85, respectively). A noticeable increase begins in 2016, with the count exceeding 500 in 2017 and reaching a peak of \num{1678} in 2020. From 2021 to 2023, the number of instances consistently exceeds \num{1500} per year, and 2024 alone contributes over 700 instances. Overall, more than 89.5\% of all instances originate from the 2018–2024 period, reflecting a strong focus on recent software development activity. 

\begin{figure}[t]
    \centering
    \includegraphics[width=0.95\linewidth]{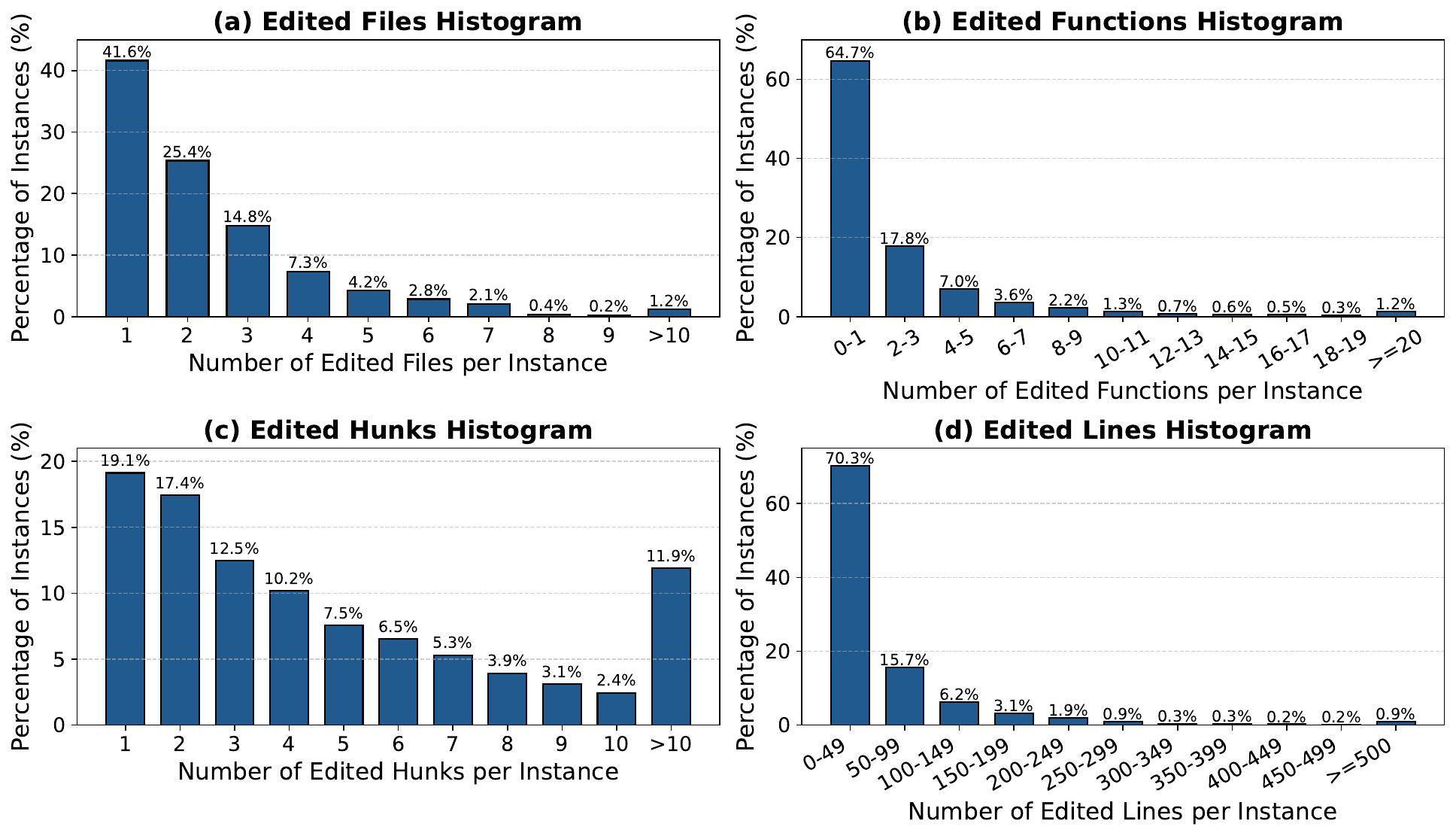}
    \vspace{-3mm}
    \caption{Statistical analysis of instance-level edits on the Skywork-SWE dataset. (a) Histogram of the number of edited files per instance. (b) Histogram of the number of edited functions per instance. (c) Histogram of the number of edited code hunks per instance. (d) Histogram of the number of edited code lines per instance.
}
    \label{fig:edited_statistical_analysis}
    \vspace{-1em}
\end{figure}

\textbf{Edit Complexity.} Skywork-SWE instances exhibit varying degrees of structural complexity in the associated patch edits. As shown in Table~\ref{tab:stats},  the average gold patch typically involves more modifications across multiple lines, files, functions, and code hunks than SWE-Gym Lite and SWE-bench Verified. Specifically, Fig.~\ref{fig:edited_statistical_analysis} (a) illustrates that 41.6\% of instances involve edits to a single file, yet over 80\% modify up to three files, demonstrating both focused and distributed complexity.  Fig.~\ref{fig:edited_statistical_analysis}(b) shows that 64.7\% of instances affect fewer than two functions, while less than 2\% involve changes to more than 15 functions. In Fig.~\ref{fig:edited_statistical_analysis}(c), nearly 50\% of instances comprise one to three hunks, whereas edits spanning more than ten hunks account for less than 12\% of cases. Finally, Fig.~\ref{fig:edited_statistical_analysis}(d) shows that 70.3\% of instances involve fewer than 50 edited lines of code, and more than 85\% remain within 100 lines.  In terms of granularity, nearly half of the instances involve one to three code hunks, and 70.3\% involve fewer than 50 edited lines, suggesting a mixture of concise and non-trivial modifications.

\textbf{Validation Strength.} Each instance is validated through unit tests, with many exhibiting multiple  cases. As shown in Table~\ref{tab:stats}, the \texttt{FAIL\_TO\_PASS} coverage per instance is more comprehensive than that of prior benchmarks. This allows for more rigorous verification of patch correctness and encourages robust evaluation of model outputs.

Overall, Skywork-SWE offers a large-scale, diverse, and structurally rich dataset that features rigorous execution-based validation and an emphasis on recent development trends, making it a valuable resource for software engineering tasks.

\subsection{Agent Trajectory Generation}
\label{sec:trajectory}

High-quality agent trajectories are crucial for agent model training. In this stage, we systematically generate, validate, and collect agent trajectories from the task instances in Skywork-SWE to offer reliable training data for our Skywork-SWE-32B model.

\textbf{Trajectory Rollout.} We leverage multiple high-performing code-focused proprietary LLMs to automatically generate agent trajectories for each task instance using OpenHands v0.32.0~\citep{wang2024openhands} code agent framework. Each trajectory is capped at a maximum of 100 rollout turns per instance.

\textbf{Trajectory Validation.} We rigorously validate the generated trajectories by applying their final patches to the corresponding task instances and executing the repository's test suite. A trajectory is considered valid if its final patch passes all tests, indicating that the underlying issue has been successfully resolved.

\textbf{Trajectory Collection.} We aggregate the validated trajectories as multi-turn supervised fine-tuning data, with only those whose final patches pass all tests in the validation step being retained. This filtering ensures consistently high data quality, directly enhancing the effectiveness of the agent model training.

Following the three aforementioned steps, we collect over \num{8000} successful agent trajectories featuring multi-turn interactions and long-context dependencies. Each trajectory is verified to correctly pass all unit tests within its corresponding runtime environment. Further details on model-specific rollout results and the trajectory composition can be found in Sec.~\ref{sec:4.3}.

\subsection{Training the Skywork-SWE Agent Model}
\label{sec:train}
After carefully filtering for format consistency, the final set of successful trajectories was reduced to 8,209 instances, which were used to train our Skywork-SWE model.
We fine-tune our Skywork-SWE agent model on the collected trajectories using supervised learning~\citep{torchtune}, with Qwen-2.5-Coder-32B-Instruct~\citep{hui2024qwen2} as the base model. Our primary goal is to validate the effectiveness of the proposed Skywork-SWE dataset and its associated runtime environments.
While reinforcement learning remains a promising direction, it requires instance-level validation within runtime images to obtain verified reward signals~\citep{cao2025skyrl}. We leave this engineering-intensive task for future work.

\begin{table}[!t]
  \centering
  \caption{Model performance across different approaches on SWE-Bench Verified. “TTS” denotes test-time scaling. “–” indicates that the exact parameter count of proprietary LLM is not publicly available. Rows corresponding to both Qwen-2.5-Coder-32B-Instruct and OpenHands are highlighted in a light cyan background. \textsuperscript{$\star$} Results are based on our own evaluation; all other results are taken from previously published work.}
  \label{tab:main}
  \resizebox{\textwidth}{!}{%
    \begin{tabular}{l r ll r}
      \toprule
      \textbf{Approach}                           & \textbf{\#Params}  & \textbf{Framework}   & \textbf{Model}                                & \textbf{Resolve Rate (\%)} \\
      \midrule
      % ---------------- Proprietary Models ----------------
      \rowcolor{gray!20}
      \multicolumn{5}{c}{\textit{\textbf{Proprietary Models}}} \\
      \cmidrule{1-5}
      OpenHands + Gemini-2.0-Flash~\citep{google2024gemini2flash}\textsuperscript{$\star$}          & –    & OpenHands      & Gemini-2.0-Flash                     & 20.0 \\
      OpenHands + OpenAI-GPT-4.1-mini~\citep{openai2024gpt41}      & –    & OpenHands      & OpenAI-GPT-4.1-mini                  & 23.6 \\
      OpenHands + Qwen-2.5-Max~\citep{qwen2024qwen25max}\textsuperscript{$\star$}              & –    & OpenHands      & Qwen-2.5-Max                         & 31.4 \\
      OpenAI-GPT-4o~\citep{openai2024gpt4o}                      & –    & Internal pipeline       & GPT-4o                               & 33.2 \\
      AutoCodeRover~\citep{zhang2024autocoderover}                      & –    & AutoCodeRover  & GPT-4o (@May~13,~2024)              & 38.4 \\
      Agentless-1.5 (GPT-4o)~\citep{xia2024agentless}              & –    & Agentless      & GPT-4o (@May~13,~2024)               & 38.8 \\
      Moatless Tools~\citep{Orwall2024Moatless}                      & –    & Moatless Tools & Claude-3-5-Sonnet-20241022           & 39.0 \\
      {OpenHands + Claude-3-5-Haiku}~\citep{anthropic2024haiku}         & –    & OpenHands      & Claude-3-5-Haiku                     & 40.6 \\
      OpenAI-o1-preview~\citep{openai2024o1pre}                   & –    & Internal pipeline       & OpenAI-o1-preview                    & 41.3 \\
      OpenHands + Claude v3.5~\citep{anthropic2024claude35sonnet} \textsuperscript{$\star$}              & –    & OpenHands  & Claude-3-5-Sonnet                            & 46.0 \\
      AutoCodeRover~v2.0~\citep{zhang2024autocoderover}                  & –    & AutoCodeRover  & Claude-3-5-Sonnet-20241022           & 46.2 \\
      OpenAI-o1~\citep{openai2024o1}                           & –    & Internal pipeline       & OpenAI-o1                            & 48.9 \\
      Agentless-1.5 (Claude)~\citep{xia2024agentless}               & –    & Agentless      & Claude-3-5-Sonnet-20241022           & 50.8 \\
      OpenHands + CodeAct~v2.1~\citep{anthropic2024claude35sonnet}             & –    & OpenHands      & Claude-3-5-Sonnet-20241022           & 53.0 \\
      OpenHands + Claude v3.7~\citep{anthropic2024claude37sonnet}\textsuperscript{$\star$}               & –    & OpenHands      & Claude-3-7-Sonnet                    & \textbf{56.0} \\
      \addlinespace
      % ---------------- Open-source Models ----------------
      \midrule
      \rowcolor{gray!20}
      \multicolumn{5}{c}{\textit{\textbf{Open-source Models}}} \\
      \cmidrule{1-5}
      SWE-Gym-7B~\citep{pan2024training}                          & 7B    & OpenHands   & Qwen-2.5-Coder-7B-Instruct           & 10.6 \\ 
      SWE-SynInfer-7B~\citep{ma2025thinking}                    & 7B    & Agentless   & Qwen-2.5-Coder-7B-Instruct           & 18.2 \\ 
      SWE-Dev-7B~\citep{wang2025swedev}                         & 7B    & OpenHands   & Qwen-2.5-Coder-7B-Instruct           & 23.4 \\ 
      \addlinespace[2pt]
      SWE-LLaMA-13B~\citep{openai2024swebench}                       & 13B   & RAG         & Code~LLaMA-13B                       &  1.2  \\  
      \addlinespace[2pt]
      Devstral~\citep{mistral2024devstral}                              & 24B   & OpenHands   & Mistral-Small-3.1-24B-Base-2503      & \textbf{46.8} \\ 
      \addlinespace[2pt]
      \rowcolor{cyan!15}
      OpenHands + Qwen~\citep{qwen2024qwen25coder}\textsuperscript{$\star$}                      & 32B   & OpenHands   & Qwen-2.5-Coder-32B-Instruct          &  6.4  \\ 
      \rowcolor{cyan!15} 
      SWE-Gym-32B~\citep{pan2024training}                        & 32B   & OpenHands   & Qwen-2.5-Coder-32B-Instruct          & 20.6 \\ 
      \rowcolor{cyan!15} 
      SWE-Dev-32B~\citep{wang2025swedev}                       & 32B   & OpenHands   & Qwen-2.5-Coder-32B-Instruct          & 36.6 \\ 
      SWE-smith-LM-32B~\citep{yang2025swe}                    & 32B   & SWE-Agent   & Qwen-2.5-Coder-32B-Instruct          & 40.2 \\ 
      \addlinespace[2pt]
      SWE-SynInfer-72B~\citep{ma2025thinking}                    & 72B   & Agentless   & Qwen-2.5-72B-Instruct                & 30.2 \\ 
      SWE-Fixer-72B~\citep{xie2025swe}                       & 72B   & Agentless   & Qwen-2.5-72B-Instruct                & 32.8 \\ 
      \addlinespace[2pt]
      OpenHands + DeepSeek-V3~\citep{deepseekai2024deepseekv3technicalreport}             & 671B  & OpenHands   & DeepSeek-V3                          & 32.4 \\ 
      OpenHands + DeepSeek-R1~\citep{guo2025deepseek}             & 671B  & OpenHands   & DeepSeek-R1                          & 34.0 \\ 
      OpenHands + DeepSeek-V3-0324~\citep{deepseekai2024deepseekv3technicalreport}        & 671B  & OpenHands   & DeepSeek-V3-0324                     & 38.8 \\ 
      \addlinespace
      % ---------------- Our Models ----------------
      \midrule
      \rowcolor{gray!20}
      \multicolumn{5}{c}{\textit{\textbf{Ours}}} \\ 
      \cmidrule{1-5}
      \rowcolor{cyan!15} 
      Skywork-SWE-32B\textsuperscript{$\star$}       & 32B   & OpenHands   & Qwen-2.5-Coder-32B-Instruct          & \textbf{38.0} \\ 
      \rowcolor{cyan!15} 
      Skywork-SWE-32B (+~TTS)\textsuperscript{$\star$}                 & 32B   & OpenHands   & Qwen-2.5-Coder-32B-Instruct          & \textbf{47.0} \\ 
      \bottomrule
    \end{tabular}%
  }
\end{table}

\begin{table}[t]
  \centering
  \caption{Summary of rollout results from various proprietary LLMs on our Skywork‐SWE dataset. Since the dataset was constructed chronologically, some trajectories were collected on earlier subsets. All rollouts were conducted using the OpenHands framework with up tp 100 interaction rounds. “–” denotes the large reasoning models that do not support the \texttt{temperature} parameter. The successful trajectories listed were further filtered down to \num{8209} instances used to train our Skywork-SWE model~(see Sec.~\ref{sec:train}).}
  \label{tab:rollout}
  \resizebox{\textwidth}{!}{
  \begin{tabular}{lccc}
    % Top thick line
    \Xhline{2\arrayrulewidth}
    \textbf{Model}            & \textbf{Temperature} & \textbf{Successful Trajectories} & \textbf{Resolve Rate (\%)} \\ 
    % Regular mid rules
    \hline
    \multirow{2}{*}{Gemini-2.0-Flash~\citep{google2024gemini2flash}}  
      & 0.0 &  482 &  5.59 \\ 
      & 1.0 &   87 &  3.63 \\ \hline
    \multirow{3}{*}{Qwen-2.5-Max~\citep{qwen2024qwen25max}}  
      & 0.0 &  717 &  8.29 \\ 
      & 0.5 &  153 &  6.38 \\ 
      & 1.0 &  520 &  8.32 \\ \hline
    Doubao-1.5-Thinking-Pro~\citep{Seed2025Doubao15pro}  
      & --  &   23 & 10.60 \\ \hline
    \multirow{2}{*}{DeepSeek-V3~\citep{deepseekai2024deepseekv3technicalreport}}  
      & 0.0 & 1071 & 12.92 \\ 
      & 1.0 &  760 & 12.18 \\ \hline
    \multirow{2}{*}{DeepSeek-V3-0324~\citep{deepseekai2024deepseekv3technicalreport}}  
      & 0.0 &  684 & 17.49 \\ 
      & 1.0 &  631 & 17.10 \\ \hline
    o3-mini~\citep{OpenAIo3mini}  
      & --  & 1908 & 15.94 \\ \hline
    GPT-4.1~\citep{openai2024gpt41}  
      & 0.0 &  142 & 18.54 \\ \hline
    Gemini-2.5-Pro~\citep{google2025gemini25pro}  
      & 0.0 &   1269 & 20.23 \\ 
    % Cumulative total row
    \hline
    \multicolumn{2}{c}{\textbf{(Cumulative) Total}} & \textbf{8447} &  \\
    % Bottom thick line
    \Xhline{2\arrayrulewidth}
  \end{tabular}
  }
\end{table}

\section{Experiments}
We structure this section as follows. We first describe the training setup, baseline methods, and evaluation benchmark in Section~\ref{sec:4.1}. We then present quantitative results on the SWE-bench Verified benchmark and analyze data scaling trends for SWE tasks in Section~\ref{sec:4.2}. Finally, we provide an in-depth experimental analysis in Section~\ref{sec:4.3}.

\subsection{Experimental Setup}
\label{sec:4.1}

\paragraph{Implementation details}
We fine-tuned the Qwen2.5-Coder-32B-Instruct~\citep{qwen2024qwen25coder} model using TorchTune framework~\citep{torchtune} on 8 NVIDIA H800 GPUs for 12 hours to obtain Skywork‐SWE‐32B model. Training was conducted using the AdamW optimizer with a weight decay of 0.01 and a cosine learning rate schedule. The Skywork‐SWE‐32B model was trained for 3 epochs on over \num{8000} multi-turn, long-context trajectories from our Skywork-SWE dataset, with a peak learning rate of 5e-5.
We evaluate our Skywork-SWE-32B model with the OpenHands agent framework with up to 100 interaction rounds under two commonly used inference settings:
\begin{itemize}[leftmargin=*]
    \item \textbf{Skywork-SWE-32B} uses a standard inference strategy with a single rollout (N=1), which is also the recommend setting for SWE-Bench.   
    \item \textbf{Skywork-SWE-32B (+~TTS)} employs test-time scaling (TTS) by generating N=8 independent rollouts per instance. The final output is selected from the trajectory with the highest score, as evaluated by the OpenHands critic model~\citep{allhands2025openhandscritic}.
\end{itemize}

We compare our Skywork-SWE-32B model on SWE-Bench Verified against a broad range of advanced code agent models, which can be categorized into \emph{proprietary} and \emph{open-source} LLMs.
\emph{Proprietary models} include OpenAI GPT~\citep{openai2024gpt4o}, Gemini~\citep{google2025gemini25pro}, Qwen~\citep{qwen2024qwen25max} and Claude~\citep{anthropic2024claude35sonnet}, as well as large reasoning models such as o3-mini~\citep{OpenAIo3mini} and Doubao~\citep{Seed2025Doubao15pro}.
These models are integrated with corresponding code agent frameworks, including OpenHands~\citep{wang2024openhands}, Moatless Tools~\citep{Orwall2024Moatless}, AutoCodeRover~\citep{zhang2024autocoderover}, and Agentless~\citep{xia2024agentless}.
\emph{Open-source models} encompass instruction-tuned and retrieval-augmented LLMs, such as Code LLaMA-13B~\citep{roziere2023code}, Qwen2.5-Coder~\citep{hui2024qwen2} at various models sizes, DeepSeek~\citep{guo2025deepseek} variants, and the Mistal Code model ~\citep{mistral2024devstral}. This broad selection enables a comprehensive comparison across both proprietary and open-source code-augmented LLMs.

\paragraph{Benchmark} We adopt SWE-bench Verified~\citep{openai2024swebench} as our evaluation benchmark and use resolve rate as our evaluation metric. For the code agent framework, we utilize version 0.32.0 of the OpenHands~\citep{wang2024openhands} code agent scaffold, configured with a maximum of 100 interaction rounds.

\subsection{Experimental Results}
\label{sec:4.2}
We present the main results in Table~\ref{tab:main} and Figure~\ref{fig:data_scaling}. Below, we highlight several key observations.

\paragraph{Skywork-SWE-32B achieve SoTA performance among open-source peers.} 
As shown in Fig.~\ref{fig:data_scaling} (Top), we report pass@1 resolved accuracy without verifiers or multiple rollouts strategies for open-source LLMs built on the OpenHands agent framework. Skywork-SWE-32B achieves 38.0\% pass@1 accuracy, surpassing previous open-source Qwen2.5-Coder-32B-based LLMs. Notably, under the same LLM backbone, our model exceeds SWE-smith-LM-32B by absolute 6.8\% points, highlighting the effectiveness of our curated training trajectories from the Skywork-SWE dataset. Furthermore, as shown in Table~\ref{tab:main}, applying test-time scaling~(TTS) techniques further improves accuracy of Skywork-SWE to 47.0\%, setting a new state-of-the-art among open-source SWE agent models at the 32B scale. These results suggest a clear data scaling law during training and demonstrate the potential of test-time scaling techniques to further enhance performance on SWE tasks.

\paragraph{Code agent frameworks matter more than model size.} 
As shown in Table~\ref{tab:main}, while larger open-source models yield slight improvements, the performance gains are relatively modest. For instance, Qwen-2.5-72B~\cite{qwen2.5} and DeepSeek-V3-671B~\cite{deepseekai2024deepseekv3technicalreport} achieves resolve rates of 30.2\% and 38.8\%, respectively. These results suggest that model size alone is not the dominant factor driving performance in software engineering tasks. Instead, task-specific high-quality training data and well-designed code agent frameworks play a more pivotal role. Among these code agent frameworks, OpenHands~\citep{wang2024openhands} stands out by consistently achieving the highest resolve rates across both proprietary and open-source models. As shown in Table~\ref{tab:main}, over half of the evaluated approaches adopt OpenHands, underscoring its effectiveness and widespread adoption for complex SWE tasks. In light of these advantages, we also adopt OpenHands as the framework for tackling SWE tasks.

\paragraph{Data scaling laws for SWE tasks.}
Fig.~\ref{fig:data_scaling} (Top) illustrates the resolve rate of Skywork-SWE-32B on SWE-bench Verified as a function of the number of training trajectories \(N_{\text{traj}}\). The performance exhibits a clear log-linear trend, consistently improving with increasing data volume~\citep{kaplan2020scaling,hoffmann2022chinchilla}.We curate high-quality, multi-turn, long-context training trajectories, each ending with a final patch that successfully passes all tests within its corresponding runtime environment. Several key data points illustrate the practical benefits of data scaling: Skywork-SWE-32B outperforms SWE-Dev-32B at $N_{\text{traj}} = 2000$, OpenHands-LM-32B-v0.1 at $6000$, and SWE-Agent-LM-32B at $8000$. These results highlight that scaling up high-quality training data can match or even surpass the performance gains achieved through more complex agent designs in SWE tasks.

\subsection{Experimental Analysis}
\label{sec:4.3}
In this section, we present detailed summary of the rollout results from various proprietary LLMs on our Skywork-SWE dataset. We then examine the impact of two test-time scaling~(TTS) evaluation strategies on model performance: Best-of N sampling and the maximum number of rollout rounds.

\paragraph{Data collection efficiency for SWE tasks is low.}
Table~\ref{tab:rollout} presents the rollout results of various models on our Skywork-SWE dataset. Since the dataset was constructed chronologically, some trajectories were collected from earlier subsets. We observe that even the most advanced proprietary LLMs achieve limited success on Skywork-SWE. Gemini-2.5-Pro yields the highest resolve rate at 20.23\%, followed by GPT-4.1 (18.54\%) and o3-mini (15.94\%). Other models perform even lower, with DeepSeek-V3 and Qwen-2.5-Max achieving 12.92\% and 8.29\% respectively with \texttt{temperature}=0 under deterministic decoding. The low performance can be attributed to the extensive diversity of GitHub repository and the large number of unit tests~(see Table~\ref{tab:stats}). Overall, executing the SWE tasks is challenging, and the efficiency of data collection is relatively low.

\begin{figure*}[!t]
    \centering
    \includegraphics[width=\linewidth]{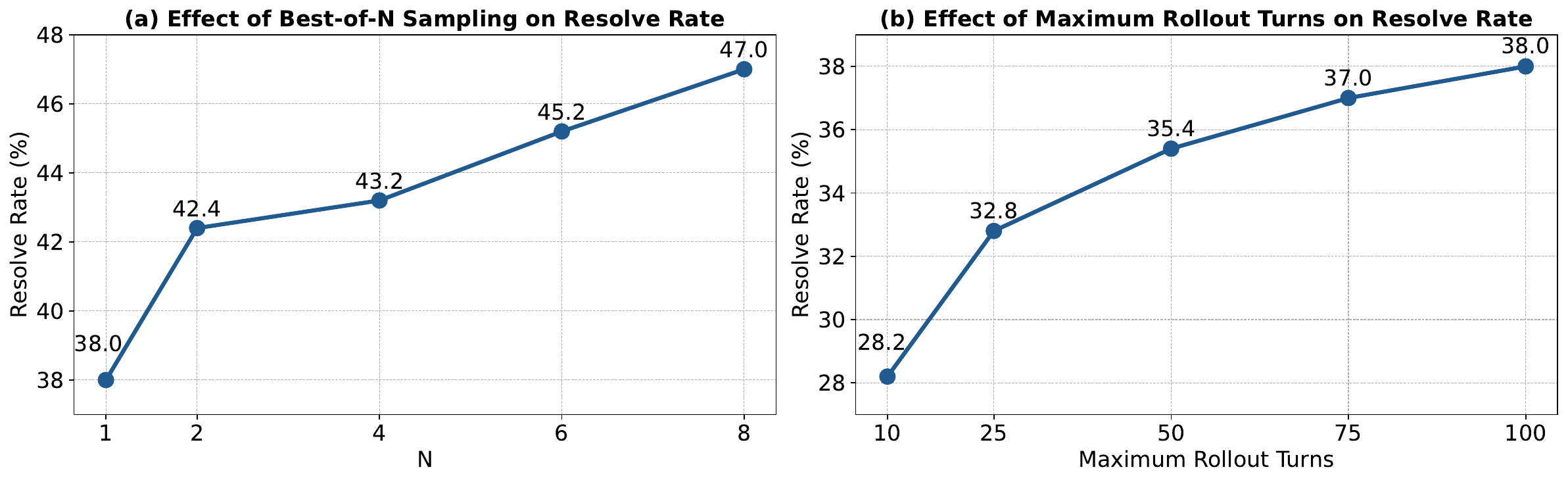}
    \vspace{-5mm}
    \caption{Resolve rate (\%) of Skywork-SWE-32B under different test-time scaling strategies.
    (a) Effect of Best-of-N Sampling with $N$ ranging from 1 to 8.
    (b) Effect of maximum rollout turns varying from 10 to 100.}
    \label{fig:turns_bon}
\end{figure*}

\paragraph{Best-of-N sampling.} \label{par:effect_bon}
The effect of varying number of N in Best-of-N sampling on the resolve rate is depicted in Figure~\ref{fig:turns_bon}(a). A clear improvement in the resolve rate is observed as the number of independent rollouts N increases. Specifically, as N is raised to 2, 4, 6, and 8, the resolve rate consistently improves, ultimately reaching 47.0\%. This trend indicates that additional rollouts help mitigate output variability, and scaling compute at inference time can significantly enhance model performance, thereby unlocking the model's inherent reasoning capabilities~\citep{brown2024large}.

\paragraph{Maximum rollout turns.} We evaluate the effect of varying the maximum number of rollout turns on the performance of Skywork-SWE-32B on SWE-bench Verified. As shown in Fig.~\ref{fig:turns_bon}(b), the resolve rate increases from 28.2\% at 10 turns to 38.0\% at 100 turns. The improvement is most notable in the early stages---from 10 to 25 turns yields a 4.6 percentage point gain---while later increases bring smaller benefits, such as a 1.0 point gain from 75 to 100 turns. SWE tasks typically require multiple rounds of interaction to resolve issues in their corresponding GitHub repositories. Therefore, LLMs benefit significantly from extended iteration scaling when the budget for test-time scaling increases with additional interaction rounds.

\section{Discussions}
Throughout the development of Skywork-SWE, we encountered several engineering challenges and setbacks. Software engineering is inherently complex and labor-intensive, particularly when it comes to tasks such as collecting GitHub repositories, configuring runtime environments, integrating code agent frameworks, and training the agent model. From our development experience, we distill a set of practical insights aimed at advancing LLM-driven software engineering in both academic research and industrial applications.

\textbf{Data Leakage Issues in Collecting GitHub Repositories.} SWE-bench Verified~\citep{openai2024swebench} includes 500 instances sourced from pull requests across 12 popular Python GitHub repositories. When curating new SWE datasets from the official PyPI~\citep{pypi} collections, it is essential to exclude repositories already included in SWE-bench Verified to prevent potential data leakage. 
Additionally, training and testing on pull request instances from the same repository can lead to inflated performance. This may occur due to partial contamination at the codebase level or greater similarity between pull requests compared to those from different repositories.
% We also empirically observe that even different pull request instances from the same repository can lead to inflated performance, as they share the same underlying codebase.

\textbf{Runtime Environment Configuration.} Each instance in an SWE task requires the corresponding runtime environment to verify whether the generated patch passes the unit tests. We adopt a default configuration~(see Sec.~\ref{sec:3.2}), which has been manually verified for reasonable coverage, to filter instances with valid environments. However, this approach inevitably results in significant data loss, as it is impossible to configure all the correct environments for different pull request instances across diverse repositories using a single unified configuration command. Considering the limited number of commonly used GitHub repositories, this high-quality data source warrants a more refined and efficient approach to environment setup. Looking ahead, developing agents capable of automatically setting up environments for the complex and diverse testing setups in real-world repositories represents a promising yet challenging direction for future work.

\textbf{Runtime Environment Reuse.} During agent trajectory generation~(see Sec.~\ref{sec:trajectory}), runtime Docker images for each task instance must be built locally to support rollout and validation. This process is highly disk-intensive; the complete set of instance-level images for the 500 SWE-bench Verified tasks requires approximately \num{1000} GB of storage. Due to limited disk capacity, images are deleted in real time after rollout and must be rebuilt for subsequent validation, resulting in redundant overhead.  
To address this, we divide the instances into mini-batches instead of completing rollout for all instances before validation. For each batch, we perform rollout, then validation, and finally delete the corresponding images to free up disk space. This improved pipeline reuses the built images for both rollout and validation, significantly reducing redundant Docker operations when storage is limited.

\textbf{Code Agent Framework.} In our work, we use version 0.32.0 of the latest OpenHands~\citep{wang2024openhands} code agent framework. We empirically observe that different versions of OpenHands exhibit variations in system prompts and execution pipelines, which can lead to substantial performance differences. Therefore, we recommend using the latest version of the OpenHands framework to evaluate SWE-bench. It is important to note that switching between OpenHands versions requires updating the corresponding SWE-bench code branch and Docker Hub namespace to ensure compatibility.

\textbf{Training the Agent Model.} Our Skywork-SWE-32B model currently supports a context length of up to \num{32768} tokens. However, we have observed that when the number of interaction rounds exceeds 50, the resulting training trajectories may surpass the 32K-token limit. To accommodate such long sequences, sequence parallelism~\citep{li2021sequence} is required in the LLM training framework when extending the model context length from 32K to 128K tokens. In future work, we plan to explore LLM training frameworks that support sequence parallelism, such as VeRL~\citep{sheng2024hybridflow} and 360-LLaMA-Factory~\citep{zhou2024360}, to enable multi-turn supervised fine-tuning with inputs up to 128K tokens.

\section{Conclusion and Future Directions}
\label{sec:conclusion}
In this report, we present an automated, execution-aware data curation pipeline used to construct our SWE dataset, comprising \num{10169} execution-validated GitHub issue–fix instances from \num{2531} repositories. Leveraging this dataset, we reveal a clear log–linear data-scaling trend on SWE-bench Verified benchmark. Our Skywork-SWE model achieves 38.0\% pass@1 accuracy on SWE-bench Verified---without verifiers or multiple rollouts---setting a new state-of-the-art among Qwen2.5-Coder-32B-based LLMs within the OpenHands agent framework. The integration of test-time scaling techniques boosts performance to 47.0\% accuracy, significantly outperforming the previous SOTA results for sub-32B parameter models. These results underscore that high-quality, execution-grounded data remain the primary bottleneck for SWE code agents, and that systematic data expansion can substantially close the gap with proprietary LLMs. We distill practical guidelines to advance LLM-driven software engineering and release the Skywork-SWE-32B checkpoint to accelerate future research.

We outline two promising directions for future work. First, existing benchmarks like SWE-bench focus almost exclusively on Python, limiting their ability to evaluate LLMs in broader software development contexts. Expanding evaluation to multiple programming languages is essential for a more comprehensive assessment of software engineering capabilities, as demonstrated by initiatives like Multi-SWE-Bench~\citep{zan2025multiswebench}.
Second, the SWE task executes and validates unit tests in the runtime environment to provide accurately verified rewards. This setup paves the way for exploring online reinforcement learning methods in the recent LLM community.
\section{Acknowledgment}
We would like to thank Xingyao Wang, the author of OpenHands code agent framework, for his support and guidance during our use of the framework. We also extend our gratitude to Jiayi Pan~(the author of SWE-Gym) and Haoran Wang~(the author of SWE-Dev) for their valuable discussions and constructive insights. Our deepest thanks goes to Mr. Yahui Zhou, the founder of Kunlun Inc., whose financial support in scaling the SWE dataset and providing access to GPU resources was indispensable for the successful completion of this study.

\clearpage
\bibliography{main}

% \appendix
% \input{sections/appendix}
% \appendix
\end{document}